\documentclass[10pt,twocolumn,letterpaper]{article}

\usepackage{3dv}
\usepackage{times}
\usepackage{epsfig}
\usepackage{graphicx}
\usepackage{caption}
\usepackage{subcaption}
\usepackage{amsmath}
\usepackage{amssymb}
\usepackage{textcomp}
\usepackage{xspace}
\usepackage{verbatim}
\usepackage{color}
\usepackage{multirow}
\usepackage[export]{adjustbox}
\usepackage{diagbox}
\usepackage{nccmath}
\usepackage{xr}
\usepackage{cuted}

\newcommand{\ignore}[1]{}
\newcommand{\fccrf}[0]{CRF\xspace}
\newcommand{\ours}[0]{SEGCloud\xspace}
\newcommand{\threedfcnn}[0]{3D-FCNN\xspace}

\newcommand{\crf}[0]{CRF\xspace}

\usepackage{enumitem}
\setitemize[0]{leftmargin=10pt}
\newenvironment{tight_itemize}{
\begin{itemize}[leftmargin=10pt]
  \setlength{\topsep}{0pt}
  \setlength{\itemsep}{2pt}
  \setlength{\parskip}{0pt}
  \setlength{\parsep}{0pt}
}{\end{itemize}}

\usepackage[pagebackref=true,breaklinks=true,letterpaper=true,colorlinks,bookmarks=false]{hyperref}

\newif\ifarxiv
\arxivtrue
\threedvfinalcopy

\def\threedvPaperID{91} 

\ifthreedvfinal\fi

\title{\ours: Semantic Segmentation of 3D Point Clouds \\
\small{\url{http://segcloud.stanford.edu}}}

\author{Lyne P. Tchapmi  \hspace{5mm} Christopher B. Choy \hspace{5mm}  Iro Armeni \hspace{5mm} JunYoung Gwak \hspace{5mm} Silvio Savarese\\
{\small lynetcha@stanford.edu \hspace{3mm} chrischoy@ai.stanford.edu \hspace{3mm} iarmeni@cs.stanford.edu \hspace{3mm} jgwak@stanford.edu \hspace{3mm} ssilvio@stanford.edu} \\
\\ Stanford University}

\begin{document}
\twocolumn[{
	\maketitle
	\vspace{-12mm}
	\begin{center}
		\includegraphics[height=6.3cm, keepaspectratio]{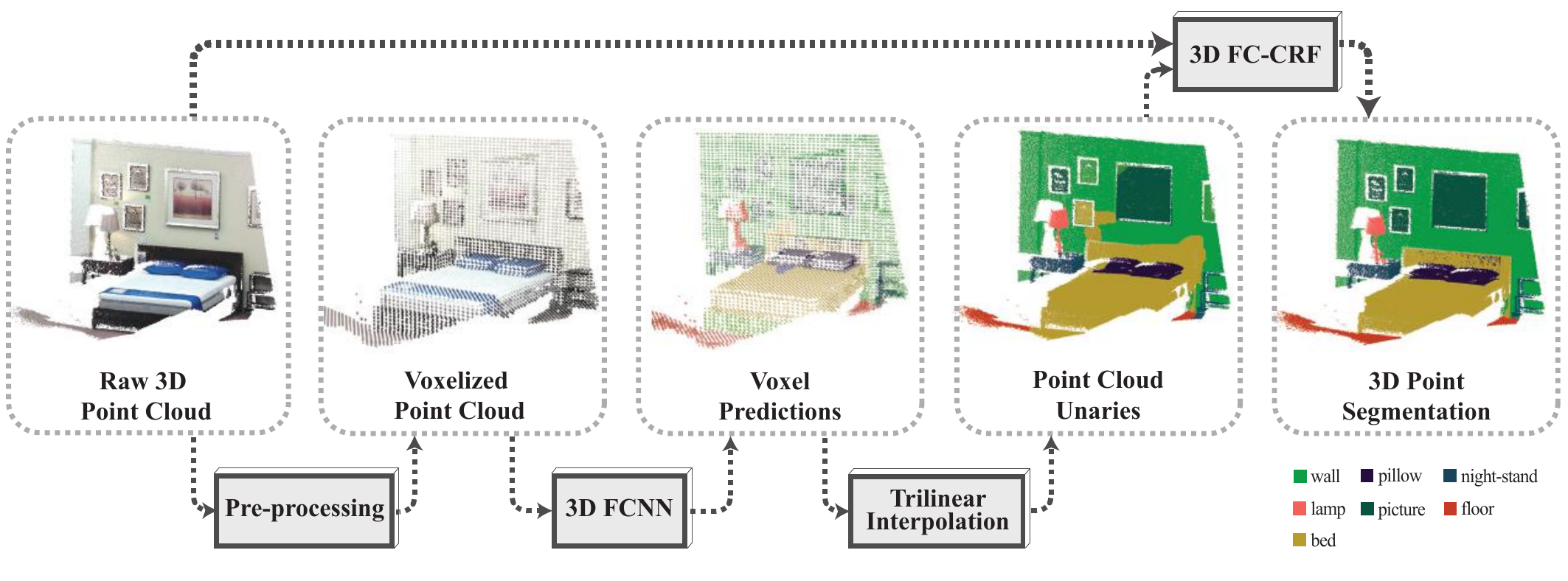}
	\end{center}
	\vspace{-6mm}
	\captionof{figure}{\small{\textbf{\ours: }A 3D point cloud is voxelized and fed through a 3D fully convolutional neural network to produce coarse downsampled voxel labels. A trilinear interpolation layer transfers this coarse output from voxels back to the original 3D Points representation. The obtained 3D point scores are used for inference in the 3D fully connected CRF to produce the final results. Our framework is trained end-to-end.}}
	\label{fig:network_overview}
    \vspace{2mm}
}]
\begin{abstract}
3D semantic scene labeling is fundamental to agents operating in the real world. In particular, labeling raw 3D point sets from sensors provides fine-grained semantics. Recent works leverage the capabilities of Neural Networks (NNs), but are limited to coarse voxel predictions and do not explicitly enforce global consistency. We present \ours, an end-to-end framework  to obtain 3D point-level segmentation that combines the advantages of NNs, trilinear interpolation(TI) and fully connected Conditional Random Fields (FC-CRF). Coarse voxel predictions from a 3D Fully Convolutional NN are transferred back to the raw 3D points via trilinear interpolation. Then the FC-CRF enforces global consistency and provides fine-grained semantics on the points. We implement the latter as a differentiable Recurrent NN to allow joint optimization. We evaluate the framework on two indoor and two outdoor 3D datasets (NYU V2, S3DIS, KITTI, Semantic3D.net), and show performance comparable or superior to the state-of-the-art on all datasets.
\end{abstract}

\section{Introduction} \label{sec:intro}
Scene understanding is a core problem in Computer Vision and is fundamental to applications such as robotics, autonomous driving, augmented reality, and the construction industry. Among various scene understanding problems, 3D semantic segmentation allows finding accurate object boundaries along with their labels in 3D space, which is useful for fine-grained tasks such as object manipulation, detailed scene modeling, etc. 

Semantic segmentation of 3D point sets or point clouds has been addressed through a variety of methods leveraging the representational power of graphical models~\cite{Koppula2011, Lu2012,Anand2013,munoz-eccv-12,hu-icra-13, Kim2013}. A common paradigm is to combine a classifier stage and a Conditional Random Field (CRF)~\cite{Lafferty2001CRF} to predict spatially consistent labels for each data point~\cite{Wolf2015,Wolf2016, Martinovic2015,conf/itsc/WangLA15,Wolf2016}. Random Forests classifiers~\cite{Breiman2001,Criminisi2013} have shown great performance on this task, however the Random Forests classifier and CRF stage are often optimized independently and put together as separate modules, which limits the information flow between them.

3D Fully Convolutional Neural Networks (\threedfcnn)~\cite{fcnn} are a strong candidate for the classifier stage in 3D Point Cloud Segmentation. However, since they require a regular grid as input, their predictions are limited to a coarse output at the voxel (grid unit) level. The final segmentation is coarse since all 3D points within a voxel are assigned the same semantic label, making the voxel size a factor limiting the overall accuracy. To obtain a fine-grained segmentation from \threedfcnn, an additional processing of the coarse \threedfcnn output is needed. We tackle this issue in our framework which is able to leverage the coarse output of a \threedfcnn and still provide a fine-grained labeling of 3D points using trilinear interpolation (TI) and CRF.

We propose an end-to-end framework that leverages the advantages of \threedfcnn, trilinear interpolation~\cite{http://bigwww.epfl.ch/publications/meijering0201.html}, and fully connected Conditional Random Fields(FC-CRF)~\cite{Lafferty2001CRF,denseCRF} to obtain fine-grained 3D Segmentation. In detail, the \threedfcnn provides class probabilities at the voxel level, which are transferred back to the raw 3D points using trilinear interpolation. We then use a Fully Connected Conditional Random Field (FC-CRF) to infer 3D point labels while ensuring spatial consistency. Transferring class probabilities to points before the \fccrf step, allows the \fccrf to use point level modalities (color, intensity, etc.) to learn a fine-grained labeling over the points, which can improve the initial coarse \threedfcnn predictions.
We use an efficient \fccrf implementation to perform the final inference. Given that each stage of our pipeline is differentiable, we are able to train the framework end-to-end using standard stochastic gradient descent.

The contributions of this work are:
\begin{tight_itemize}
    \item{We propose to combine the inference capabilities of Fully Convolutional Neural Networks with  the fine-grained representation of 3D Point Clouds using TI and CRF.}
    \item{We train the voxel-level \threedfcnn and point-level CRF jointly and end-to-end by connecting them via Trilinear interpolation enabling segmentation in the original 3D points space.}
\end{tight_itemize}

Our framework can handle 3D point clouds from various sources (laser scanners, RGB-D sensors,  etc.), and we demonstrate state-of-the art performance on indoor and outdoor, partial and fully reconstructed 3D scenes, namely on NYU V2\cite{Silberman:ECCV12}, Stanford Large-Scale 3D Indoor Spaces Dataset (S3DIS)\cite{Armeni2016}, KITTI\cite{Geiger2013IJRR,Geiger:2012:WRA:2354409.2354978}, and the Semantic3D.net benchmark for outdoor scenes\cite{l3d}.

\section{Related Work}
\begin{figure*}[t]
\centering
\vspace{-10mm}

\includegraphics[width=.99\linewidth]{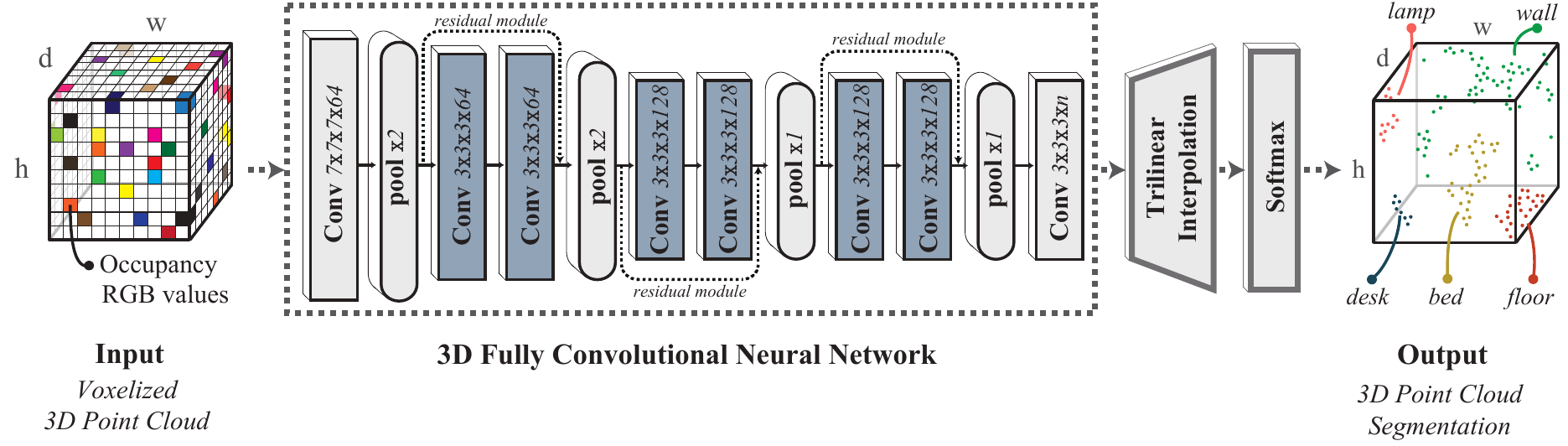}

\caption{\small{\textbf{Network architecture}: The \threedfcnn is made of 3 residual layers sandwiched between 2 convolutional layers. Max Pooling in the early stages of the network yields a 4X downsampling.}}
\label{fig:network_architecture}
\vspace{-5mm}
\end{figure*}

In this section, we present related works with respect to three main aspects of our framework: neural networks for 3D data, graphical models for 3D Segmentation and works that explore the combination of Convolutional Neural Networks (CNN) and CRF. Other techniques have been employed for 3D Scene Segmentation~\cite{Cohen, Aijazi, Lalonde_2006_5611} but we focus mainly on the ones related to the above topics.

\textbf{Neural Networks for 3D Data:} 3D Neural Networks have been extensively used for 3D object and parts recognition~\cite{DeepSlidingShapes, qi2016volumetric, maturana_iros_2015, Guo2015, DBLP:journals/corr/QiSMG16, Garcia}, understanding object shape priors, as well as generating and reconstructing objects~\cite{Yan2016, 3dgan, fan2016, 3dinterpreter, choy20163d}. Recent works have started exploring the use of Neural Networks for 3D Semantic Segmentation~\cite{DBLP:journals/corr/QiSMG16, dai2017scannet, J.Huang2016}. Qi ~\etal.~\cite{DBLP:journals/corr/QiSMG16} propose a Multilayer Perceptron (MLP) architecture that extracts a global feature vector from a 3D point cloud of 1$m^3$ physical size and processes each point using the extracted feature vector and additional point level transformations. Their method operates at the point level and thus inherently provides a fine-grained segmentation. It works well for indoor semantic scene understanding, although there is no evidence that it scales to larger input dimensions without additional training or adaptation required. Huang~\etal. \cite{J.Huang2016} present a \threedfcnn for 3D semantic segmentation which produces coarse voxel-level segmentation. Dai~\etal.~\cite{dai2017scannet} also propose a fully convolutional architecture, but they make a single prediction for all voxels in the same voxel grid column. This makes the wrong assumption that a voxel grid column contains 3D points with the same object label. All the aforementioned methods are limited by the fact that they do not explicitly enforce spatial consistency between neighboring points predictions and/or provide a coarse labeling of the 3D data. In contrast, our method makes fine-grained predictions for each point in the 3D input, explicitly enforces spatial consistency and models class interactions through a \fccrf. Also, in contrast to \cite{DBLP:journals/corr/QiSMG16}, we readily scale to larger and arbitrarily sized inputs, since our classifier stage is fully convolutional.

\textbf{Graphical Models for 3D Segmentation:} Our framework builds on top of a long line of works combining graphical models(~\cite{Taskar2004, Taskar2003,Lafferty2001CRF,Felzenszwalb2004,Ladicky2014}) and highly engineered classifiers. Early works on 3D Semantic Segmentation formulate the problem as a graphical model built on top of a set of features. Such models have been used in several works to capture contextual relationships based on various features and cues such as appearance, shape, and geometry. These models are shown to work well for this task~\cite{Munoz2009,Munoz2008, Koppula2011, Shapovalov2010, Lu2012,Anand2013,munoz-eccv-12}.

A common paradigm in 3D semantic segmentation combines a classifier stage and a Conditional Random Field to impose smoothness and consistency~\cite{Wolf2015,Wolf2016, Martinovic2015,conf/itsc/WangLA15,Wolf2016}. Random Forests~\cite{Breiman2001,Criminisi2013} are a popular choice of classifier in this paradigm and in 3D Segmentation in general~\cite{Zhang2015,Dohan2015,Chehata09airbornelidar,Brostow2008,Nan2012,Weinmann2014}; they use hand-crafted features to robustly provide class scores for voxels, oversegments or 3D Points. In~\cite{Martinovic2015}, the spin image descriptor is used as a feature, while~\cite{Wolf2015} uses a 14-dimensional feature vector based on geometry and appearance. Hackel~\etal.~\cite{Hackel2016} also define a custom set of features aimed at capturing geometry, appearance and location. In these works, the Random Forests output is used as unary potentials (class scores) for a CRF whose parameters are learned independently. The CRF then leverages the confidence provided by the classifier, as well as similarity between an additional set of features, to perform the final inference. In contrast to these methods, our framework uses a \threedfcnn which can learn higher dimensional features and provide strong unaries for each data point. Moreover, our \fccrf is implemented as a fully differentiable Recurrent Neural Network, similar to \cite{Zheng2015}. This allows the \threedfcnn and \fccrf to be trained end-to-end, and enables information flow from the CRF to the CNN classification stage.

\textbf{Joint CNN + CRF:} Combining 3D CNN and 3D CRF has been previously proposed for the task of lesion segmentation in 3D medical scans. Kamnitsas~\etal.~\cite{Kamnitsas2017} propose a multi-scale 3D CNN with a \fccrf to classify 4 types of lesions from healthy brain tissues. The method consists of two modules that are not trained end-to-end: a 2-stream architecture operating at 2 different scan resolutions and a \fccrf. In the \fccrf training stage, the authors reduce the problem to a 2-class segmentation task in order to find parameters for the CRF that can improve segmentation accuracy. 

Joint end-to-end training of CNN and CRF was first demonstrated by~\cite{Zheng2015} in the context of image semantic segmentation, where the CRF is implemented as a differentiable Recurrent Neural Network (RNN). The combination of CNN and CRF trained in an end-to-end fashion demonstrated state-of-the-art accuracy for semantic segmentation in images. In~\cite{Zheng2015} and other related works~\cite{fcnn, Deeplab}, the CNN has a final upsampling layer with learned weights which allows to obtain pixel level unaries before the CRF stage. Our work follows a similar thrust by defining the CRF as an RNN and using a trilinear interpolation layer to transfer the coarse output of the \threedfcnn to individual 3D points before the CRF stage. In contrast to~\cite{Kamnitsas2017}, our framework is a single stream architecture which jointly optimizes the 3D CNN and \crf, targets the domain of 3D Scene Point Clouds, and is able to handle a large number of classes both at the CNN and CRF stage. Unlike~\cite{Zheng2015,fcnn, Deeplab}, we choose to use deterministic interpolation weights that take into account the metric distance between a 3D point and its neighboring voxel centers (Section \ref{sec:trilinear}). Our approach reduces the number of parameters to be learned, and we find it to work well in practice. We show that the combination of jointly trained \threedfcnn and \crf with TI consistently performs better than a stand alone \threedfcnn.

In summary, our work differs from previous works in the design of an end-to-end deep learning framework for fine-grained 3D semantic segmentation, the use of deterministic trilinear interpolation to obtain point-level segmentation, and the use of a jointly trained \fccrf to enforce spatial consistency. The rest of the paper is organized as follows. Sections \ref{sec:framework} and \ref{sec:joint} present the components of our end-to-end framework and Section \ref{sec:implementation} provides implementation details. Section \ref{sec:experiments} presents our experiments including datasets (\ref{sec:datasets}), benchmark results (\ref{sec:results}), and system analysis (\ref{sec:system}). Section \ref{sec:conclusion} concludes with a summary of the presented results.

\section{\ours Framework} \label{sec:framework}
An overview of the \ours pipeline is shown in Figure~\ref{fig:network_overview}. In the first stage of our pipeline, the 3D data is voxelized and the resulting 3D grid is processed by a 3D fully convolutional neural network (\threedfcnn)\footnote{Depending on the type of 3D data a pre-processing step of converting it to a 3D point cloud representation might be necessary.}. The \threedfcnn down-samples the input volume and produces probability distributions over the set of classes for each down-sampled voxel (Section~\ref{sec:threedcnn}). The next stage is a trilinear interpolation layer which interpolates class scores from down-sampled voxels to 3D points (Section~\ref{sec:trilinear}). Finally, inference is performed using a \fccrf which combines the original 3D points features with interpolated scores to produce fine-grained class distributions over the point set (Section~\ref{sec:crf}). Our entire pipeline is jointly optimized and the \fccrf inference and joint optimization processes are presented in Section~\ref{sec:joint}.

\subsection{3D Fully Convolutional Neural Network} \label{sec:threedcnn}

Our framework uses a \threedfcnn to learn a representation suitable for semantic segmentation. Moreover, the fully convolutional network reduces the computational overhead needed to generate predictions for each voxel by sharing computations~\cite{Long2015}. In the next section, we describe how we represent 3D point clouds as an input to the \threedfcnn.

\noindent\textbf{\threedfcnn data representation: }Given that the  \threedfcnn input should be in the form of a voxel grid, we convert 3D point clouds as follows. Each data point is a 3D observation $o_i$, that consists of the 3D position $p_i$ and other available modalities, such as the color intensity $I_i$ and sensor intensity $S_i$. We place the 3D observations $\mathbf{O}=\{o_i\}$ in a metric space so that the convolution kernels can learn the scale of objects. This process is usually handled in most 3D sensors. Then we define a regular 3D grid that encompasses the 3D observations. We denote each cell in the 3D grid as a voxel $v_i$ and for simplicity, each cell is a cube with length $V = 5cm$. Most of the space in the 3D input is empty and has no associated features. To characterize this, we use a channel to denote the occupancy as a binary value (zero or one). We use additional channels to represent other modalities. For instance, three channels are used for RGB color, and one channel is used for sensor intensity when available.

\noindent\textbf{Architecture:} Our \threedfcnn architecture is illustrated in Figure~\ref{fig:network_architecture}. We use 3 residual modules~\cite{ResNet} sandwiched between 2 convolutional layers, as well as 2 destructive pooling layers in the early stages of the architecture to down-sample the grid, and 2 non-destructive ones towards the end. The early down-sampling gives us less memory footprint. The entire framework is fully convolutional and can handle arbitrarily sized inputs. For each voxel $v_i$, the \threedfcnn outputs scores(logits) $L_i$ associated with a probability distribution  $q_i$ over labels. The resulting scores are transferred to the raw 3D points via trilinear interpolation.

\subsection{3D Trilinear Interpolation} \label{sec:trilinear}

\begin{figure}
    \centering
    \includegraphics[width=.99\linewidth]{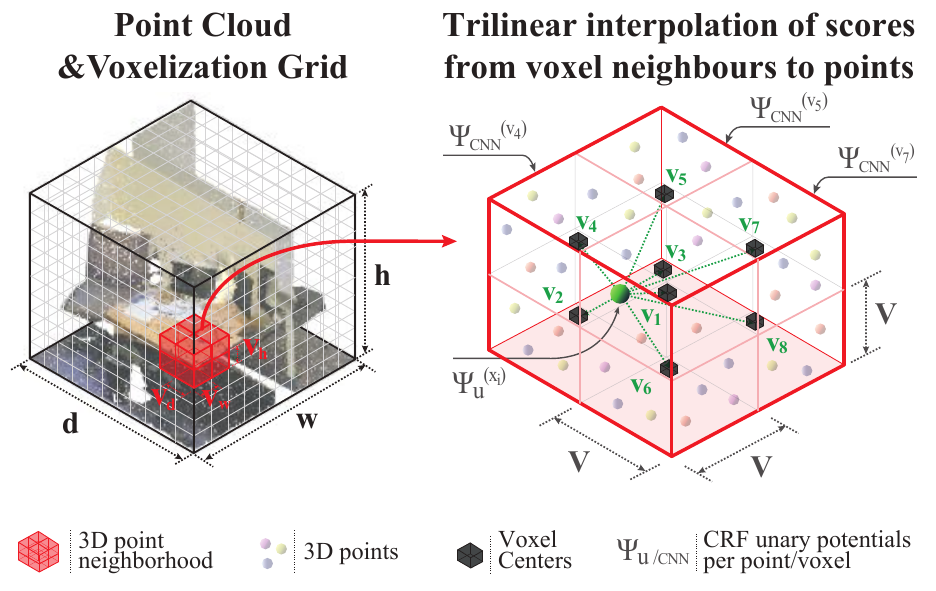}
    \caption{\small{\textbf{Trilinear interpolation of class scores from voxels to points}: Each point's score is computed as the weighted sum of the scores from its 8 spatially closest voxel centers.}}
    \label{fig:CRF}
\end{figure}

The process of voxelization and subsequent down-sampling in the \threedfcnn converts our data representation to a coarse 3D grid which limits the resolution of semantic labeling at the CRF stage (to 20 cm in our case). Running the CRF on such coarse voxels results in a coarse segmentation. One option to avoid this information loss is to increase the resolution of the voxel grid (\ie decrease the voxel size) and/or remove the destructive pooling layers, and run the CRF directly on the fine-grained voxels. However, this quickly runs into computational and memory constraints, since for given 3D data dimensions, the memory requirement of the \threedfcnn grows cubically with the resolution of the grid. Also, for a given \threedfcnn architecture, the receptive field decreases as the resolution of the grid increases, which can reduce performance due to having less context available during inference(see \cite{segcloudsuppl}). 

We therefore dismiss a voxel-based CRF approach and resort to running \fccrf inference using the raw 3D points as nodes. In this way, the CRF can leverage both the \threedfcnn output and the fine-grained modalities of the input 3D points to generate accurate predictions that capture scene and object boundaries in detail. We achieve this using trilinear interpolation to transfer the voxel-level predictions from the \threedfcnn to the raw 3D points as illustrated in Figure~\ref{fig:CRF}. Specifically, for each point, $o_i = \{p_i, I_i, S_i\}$, we define a random variable $x_i$ that denotes the semantic class, and the scores(logits) $L_i$ associated with the distribution of $x_i$ are defined as a weighted sum of scores $L_{i,n}(x_{i,n})$ of its 8 spatially closest voxels $v_{i,n}$, $n\in\{1,..., 8\}$ whose centers are $(p_{i,n}^x, p_{i,n}^y, p_{i,n}^z)$ as follows:
\vspace{-2mm}
\begin{ceqn}\label{eq:unary}
\begin{align} 
    \psi_u(x_i = l) & =  L_i(x_i = l) =  \sum_{n=1}^{8}w_{i,n} L_{i,n}(x_{i,n} = l) \\
    w_{i,n} & = \prod_{s \in \{x, y, z\}} \left( 1 - |p_i^s-p_{i,n}^s|/V \right) \nonumber
\end{align}
\end{ceqn}

where $V$ is the voxel size. During back propagation, we use the same trilinear interpolation weights $w_{i,n}$ to splat the gradients from the \fccrf to the \threedfcnn. The obtained point level scores are then used as unaries in the CRF.

\subsection{3D Fully Connected Conditional Random Field} \label{sec:crf}

The energy function of a \fccrf consists of a set of unary and pairwise potential energy terms. The unary potentials are a proxy for the initial probability distribution across semantic classes and the pairwise potentials enforce smoothness and consistency between predictions.
The energy of the \fccrf is defined as,

\begin{equation} \label{eq:crf}
    E(x) = \sum_{i}\psi_u(x_{i}) + \sum_{i<j}\psi_p(x_i,x_j)
\end{equation}

where $\psi_u$ denotes the unary potential which is defined in Equation~\ref{eq:unary} and $\psi_p$ denotes the pairwise potential. Note that all nodes in the \fccrf are connected with each other through the pairwise potentials. We use the Gaussian kernels from~\cite{denseCRF} for the pairwise potentials,
\vspace{-2mm}
\begin{align}
\begin{split}
    \psi_p(x_i,x_j) & = \mu(x_i,x_j) \left[ w_s \exp{ \left( - \frac{|p_i - p_j|^2}{2\theta^2_\gamma} \right) } \right. \\
    & \left. + w_b \exp{ \left(- \frac{|p_i - p_j|^2}{2\theta^2_\alpha} - \frac{|I_i - I_j|^2}{2\theta^2_\beta} \right)} \right]
\end{split}
\end{align}

where $w_b$ and $w_s$ are the weights of the bilateral and spatial kernel respectively, $\mu$ is the label compatibility score, and $\theta_\alpha, \theta_\beta, \theta_\gamma$ are the kernels' bandwidth parameters. When RGB information is not available, we only use the spatial kernel. Using Gaussian kernels enables fast variational inference and learning through a series of convolutions on a permutohedral lattice~\cite{permutohedral_convolution} (Section~\ref{sec:inference}). 

\section{\fccrf Inference and Joint Optimization} \label{sec:joint}
Exact energy minimization in \fccrf is intractable, therefore we rely on a variational inference method which allows us to jointly optimize both the \fccrf and \threedfcnn~\cite{Zheng2015,denseCRF}. The output after the \fccrf energy minimization gives us fine-grained predictions for each 3D point that takes smoothness and consistency into account. Given the final output of the \fccrf, we follow the convention and use the distance between the prediction and ground truth semantic labels as a loss function and minimize it.

\noindent\textbf{\fccrf Inference:} \label{sec:inference}
The \fccrf with Gaussian potential has a special structure that allows fast and efficient inference.  Kr\"{a}henb\"{u}hl~\etal. \cite{denseCRF} presented an approximate inference method which assumes independence between semantic label distributions $Q(\mathbf{X}) = \prod\limits_i
Q_i(x_i)$, and derived the update equation:
\vspace{-2mm}
\begin{align}
\begin{split}
    Q_i^+ & (x_i = l)  = \frac{1}{Z_i} \exp\Big\{ - \psi_u(x_i) \\
        & - \sum_{l' \in \mathcal{L}}\mu(l, l') \sum_{m=1}^K w^{(m)} \sum_{j \neq i} k^{(m)} (f_i, f_j) Q_j(l')\Big\}
\end{split}
\end{align}

The above update equation can be implemented using simple convolutions, sums and softmax as shown by Zheng~\etal.~\cite{Zheng2015}, who implemented CRF inference and learning as a Recurrent Neural Network (RNN), named CRF-RNN. CRF-RNN can be trained within a standard CNN framework, so we follow the same procedure to define our 3D \fccrf as an RNN for inference and learning. This formulation allows us to integrate the \fccrf within our \threedfcnn framework for joint training.

\noindent\textbf{Loss:} \label{sec:loss}
Once we minimize the energy of the \fccrf in Equation~\ref{eq:crf}, we obtain the final prediction distribution of the semantic class $x_i$ on each 3D observation $o_i$. Denoting the ground truth discrete label of the observation $o_i$ as $y_i$, we follow the convention and define our loss function as the distance between a final prediction distribution and the ground truth distribution using KL divergence:

\begin{equation}
    L(\mathbf{x}, \mathbf{y}) = \frac{1}{N} \sum_{i=1}^N E_{y_i}[- \log{p(x_i)}]
\end{equation}

where $N$ is the number of observations. Since the entropy of $y_i$ is a constant with respect to all parameters, we do not include it in the loss function equation.

\section{Implementation Details} \label{sec:implementation}
We implemented the \ours framework using the Caffe neural network library~\cite{jia2014caffe}\footnote{We use~\cite{video-caffe} that supports 3D convolution.}. Within theCaf fe framework, we adapted the bilinear interpolation of \cite{bilinear} and implemented trilinear interpolation as a neural network layer. All computations within the \threedfcnn, trilinear interpolation layer, and \fccrf are done on a Graphical Processing Unit (GPU). For \fccrf inference, we adapt the RNN implementation of Zheng ~\etal.~\cite{Zheng2015} to 3D point clouds.

To address the lack of data in some datasets and make the network robust, we applied various data augmentation techniques such as random color augmentation, rotation along the upright direction, and points sub-sampling. The above random transformations and sub-sampling allow to increase the effective size of each dataset by at least an order of magnitude, and can help the network build invariance to rotation/viewpoint changes, as well as reduced and varying context (see \cite{segcloudsuppl}).

Training is performed in a 2-step process, similar to~\cite{Zheng2015} (see Figure \ref{fig:training_stages}). In the first stage, we train the \threedfcnn in isolation via trilinear interpolation  for $200$ epochs.

In the second stage, we jointly train the \threedfcnn and the \fccrf end-to-end (both modules connected through the trilinear interpolation layer).
The approximate variational inference method we used for the \fccrf~\cite{denseCRF} approximates convolution in a permutohedral grid whose size depends on the bandwidth parameters $\theta_\alpha, \theta_\beta, \theta_\gamma$.  We fixed $\theta_\gamma$ at 5cm, $\theta_\beta$ at 11 and used a grid search with small perturbation on a validation set to find the optimal $\theta_\alpha$ (see \cite{segcloudsuppl}).

\begin{table*}[ht]
    \centering
    \vspace{-6mm}
	\caption{\small{\textbf{Results on the Semantic3D.net Benchmark (\textit{reduced-8} challenge)}}}
	\vspace{-2mm}
	\resizebox{\textwidth}{!}{
	\begin{tabular}{c|cccccccc|cc}
	 \multirow{2}{*}{\textbf{Method}} & \small{man-made} & \small{natural} & \small{high} & \small{low} & \multirow{2}{*}{\small{buildings}} & \small{hard}  & \small{scanning}  & \multirow{2}{*}{\small{cars}} & \multirow{2}{*}{\small{mIOU}} & \multirow{2}{*}{\small{mAcc}\footnote{We downloaded confusion matrices from the benchmark website to compute the mean accuracy.}}\\
	 & \small{terrain}  & \small{terrain} & \small{vegetation} & \small{vegetation} &  & \small{scape} & \small{artefacts} &  & & \\
	 \hline
	TMLC-MSR~\cite{Hackel2016} & \textbf{89.80} & 74.50 & 53.70 & 26.80 & 88.80 & 18.90 & 36.40 & 44.70 & 54.20 & 68.95\\
	DeePr3SS~\cite{1705.03428}  & 85.60 & \textbf{83.20} & 74.20 & 32.40 & 89.70 &  18.50 &  25.10 & 59.20 & 58.50 & 88.90\\
	SnapNet~\cite{snapnet}     & 82.00 & 77.30 & 79.70 & 22.90 & \textbf{91.10} & 18.40 & \textbf{37.30} & \textbf{64.40} & 59.10 & 70.80 \\
	\threedfcnn-TI(Ours)    & 84.00 & 71.10 & 77.00 & 31.80 & 89.90 & 27.70 & 25.20 & 59.00 & 58.20 & 69.86\\
	\ours(Ours)      & 83.90 & 66.00 & \textbf{86.00} & \textbf{40.50} & \textbf{91.10} & \textbf{30.90} & 27.50 & 64.30 & \textbf{61.30} & \textbf{73.08} \\	
    \end{tabular}}
    \label{tab:large_scale_outdoors}
    \vspace{-2mm}
\end{table*}
{\small
\tabcolsep= 1mm
\begin{table*}[ht]
    \centering
    \caption{\small{\textbf{Results on the  Large-Scale 3D Indoor Spaces Dataset (S3DIS)}}}
    \vspace{-2mm}
    \resizebox{\textwidth}{!}{
    \begin{tabular}{r|ccccccccccccc|cc}
        \small{\textbf{Method}} & \small{ceiling} & \small{floor} & \small{wall} & \small{beam} & \small{column} & \small{window} & \small{door} & \small{chair} & \small{table} & \small{bookcase} & \small{sofa} & \small{board} & \small{clutter} & \small{\textbf{mIOU}} & \small{\textbf{mAcc}}\\ \hline
        PointNet \cite{DBLP:journals/corr/QiSMG16} & 88.80 & \textbf{97.33} & 69.80 & \textbf{0.05} & 3.92 & \textbf{46.26} & 10.76 & 52.61 & 58.93 & 40.28 & 5.85 & \textbf{26.38} & 33.22 & 41.09 & 48.98\\
        \threedfcnn-TI(Ours) & \textbf{90.17} & 96.48 & \textbf{70.16} & 0.00 & 11.40 & 33.36 & 21.12 & \textbf{76.12} & 70.07 & 57.89 & 37.46 & 11.16 & \textbf{41.61} & 47.46 & 54.91\\
        \ours(Ours) & 90.06 & 96.05 & 69.86 & 0.00 & \textbf{18.37} & 38.35 & \textbf{23.12} & 75.89 & \textbf{70.40} & \textbf{58.42} & \textbf{40.88} & 12.96 & 41.60 & \textbf{48.92} & \textbf{57.35}\\
    \end{tabular}}
    \label{tab:S3DIS}
    \vspace{-4mm}
\end{table*}}
\begin{table*}[ht]
    \tabcolsep= 0.5mm
	\caption{\small{\textbf{Results on the NYUV2 dataset}}}
	\resizebox{\textwidth}{!}{
	\begin{tabular}{c|ccccccccccccc|ccc}
	\textbf{Method} & \small{Bed} & \small{Objects} & \small{Chair} & \small{Furniture} & \small{Ceiling} & \small{Floor}  & \small{Deco.}  & \small{Sofa} & \small{Table} & \small{Wall} & \small{Window} & \small{Booksh.} & \small{TV} & \textbf{mIOU} & \textbf{mAcc}&  \textbf{glob Acc}\\
	 \hline
	 Couprie et al.~\cite{couprie2013} & 38.1 & 8.7 & 34.1 & 42.4 & 62.6 & 87.3 & 40.4 & 24.6 & 10.2 & 86.1 & 15.9 & 13.7 & 6.0  & - & \small{36.2} & \small{52.4}\\
	 Wang et al.~\cite{Wang2014} & 47.6 & 12.4 & 23.5 & 16.7 & 68.1 & 84.1 & 26.4 & 39.1 & 35.4 & 65.9 & 52.2 & 45.0 & 32.4  & - & \small{42.2} & \small{-}\\
	 Hermans et al.~\cite{Hermans2014} & 68.4 & 8.6 & 41.9 & 37.1 & \textbf{83.4} & 91.5 & 35.8 & 28.5 & 27.7 & 71.8 & 46.1 & 45.4 & \textbf{38.4}  & - & \small{48.0} & \small{54.2}\\
	Wolf et al.~\cite{Wolf2016} & 74.56 & 17.62 & 62.16 & 47.85 & 82.42 & \textbf{98.72} & 26.36 & \textbf{69.38} & \textbf{48.57} & 83.65 & 25.56 & \textbf{54.92} & 31.05 & 39.51 & \small{55.6$\pm$0.2} & \small{64.9$\pm$0.3}\\
	\threedfcnn-TI(Ours)  & 69.3 & \textbf{40.26} & \textbf{64.34} & \textbf{64.41} & 73.05 & 95.55 & 21.15 & 55.51 & 45.09 & \textbf{84.96} & 20.76 & 42.24 & 23.95 & 42.13 & 53.9 & \textbf{67.38}\\
	\ours(Ours)  & \textbf{75.06} & 39.28 & 62.92 & 61.8 & 69.16 & 95.21 & \textbf{34.38} & 62.78 & 45.78 & 78.89 & \textbf{26.35} & 53.46 & 28.5 & \textbf{43.45} & \textbf{56.43} & 66.82\\
    \end{tabular}}
    \label{tab:NYUV2}
\end{table*}
{\small
\tabcolsep= 1mm
\begin{table*}[ht]
    \centering
    \caption{\small{\textbf{Results on the KITTI dataset.}}}
    \begin{tabular}{r|cccccccccc|cc}
        \small{\textbf{Method}} & \small{building} & \small{sky} & \small{road} & \small{vegetation} & \small{sidewalk} & \small{car} & \small{pedestrian} & \small{cyclist} & \small{signage} & \small{fence} & \small{\textbf{mIOU}} & \small{\textbf{mAcc}} \\ \hline
        Zhang~\etal.~\cite{Zhang2015} & \textbf{86.90} & - & 89.20 & 55.00 & 26.20 & 50.0 & 49.00 & \textbf{19.3} & \textbf{51.7} & \textbf{21.1} & - & \textbf{49.80}\\
        \threedfcnn-TI(Ours) & 85.83 & - & \textbf{90.57} & \textbf{70.50} & 25.56 & 65.68 & 46.35 & 7.78 & 28.40 & 4.51 & 35.65 & 47.24 \\
        \ours(Ours) & 85.86 & - & 88.84 & 68.73 & \textbf{29.74} & \textbf{67.51} & \textbf{53.52} & 7.27 & 39.62 & 4.05 & \textbf{36.78} & 49.46\\
    \end{tabular}
    \label{tab:kitti}
\end{table*}}

\section{Experiments} \label{sec:experiments}
In this section, we evaluate our framework on various 3D datasets and analyze the performance of key components.

\subsection{Datasets}
\label{sec:datasets}

Several 3D Scene datasets have been made available to the research community ~\cite{Riemenschneider2014,armeni2017,Armeni2016, scenenn-3dv16, Song2015,Xiao2013,Silberman:ECCV12,dai2017scannet,kitchen2016, catdata2013}. We chose four of them so that they cover indoor and outdoor, partial and fully reconstructed, as well as small, medium and large scale point clouds. For our evaluation, we favor those for which previous 3D Semantic Segmentation works exist, with replicable experimental setups for comparison. We choose baselines so that they are representative of the main research thrusts and topics related to our method (\ie, Neural Networks, Random Forests, and CRFs). The datasets we chose for evaluation are the Semantic3D.net Benchmark~\cite{l3d},
the Stanford Large-Scale 3D Indoor Spaces Dataset (S3DIS)~\cite{Armeni2016},  KITTI~\cite{Geiger2013IJRR,Geiger:2012:WRA:2354409.2354978}, and NYU V2~\cite{Silberman:ECCV12}.
The datasets showcase a wide range of sizes from the smallest
KITTI dataset with 12 million
training points, to the largest Semantic3D.net with $1.9$ billion training points~\footnote{This excludes the validation set in our data split}(details in \cite{segcloudsuppl}). We evaluate our method on each dataset and provide a comparison against the state-of-the-art.

\begin{figure}[t]
\centering
\includegraphics[width=.90\linewidth]{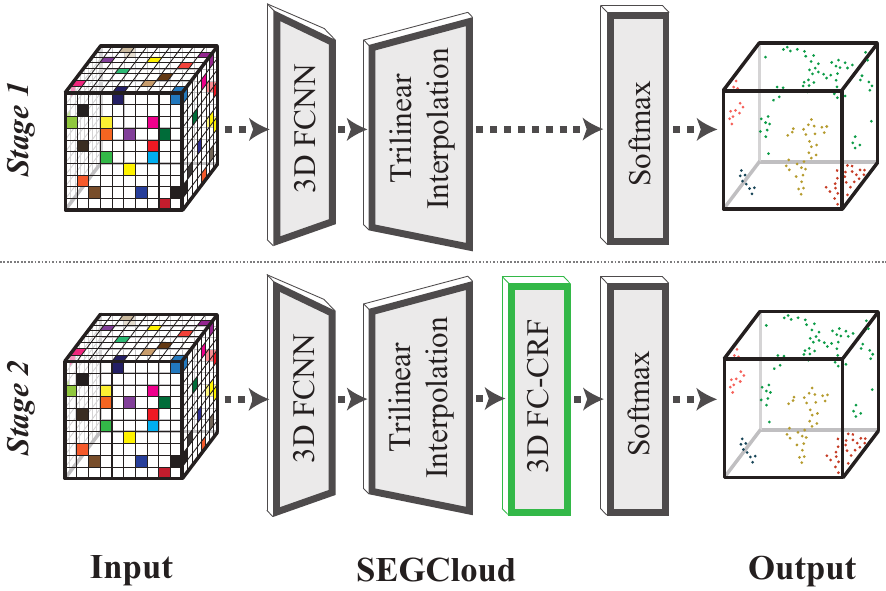}
\caption{\small{We follow a 2-stage training by first optimizing over the point-level unary potentials (no CRF) and then over the joint framework for point-level fine-grained labeling.}}
\label{fig:training_stages}
\vspace{-4mm}
\end{figure}

\subsection{Results}
\label{sec:results}

We present quantitative and qualitative results for each of the datasets introduced above. We compare against the state-of-the-art, and perform an ablation study to showcase the benefits of the \fccrf. The metrics reported are mean IOU and mean Accuracy across classes unless otherwise stated.\\

\begin{figure*}
\centering
\vspace{-4mm}

\includegraphics[width=.99\textwidth]{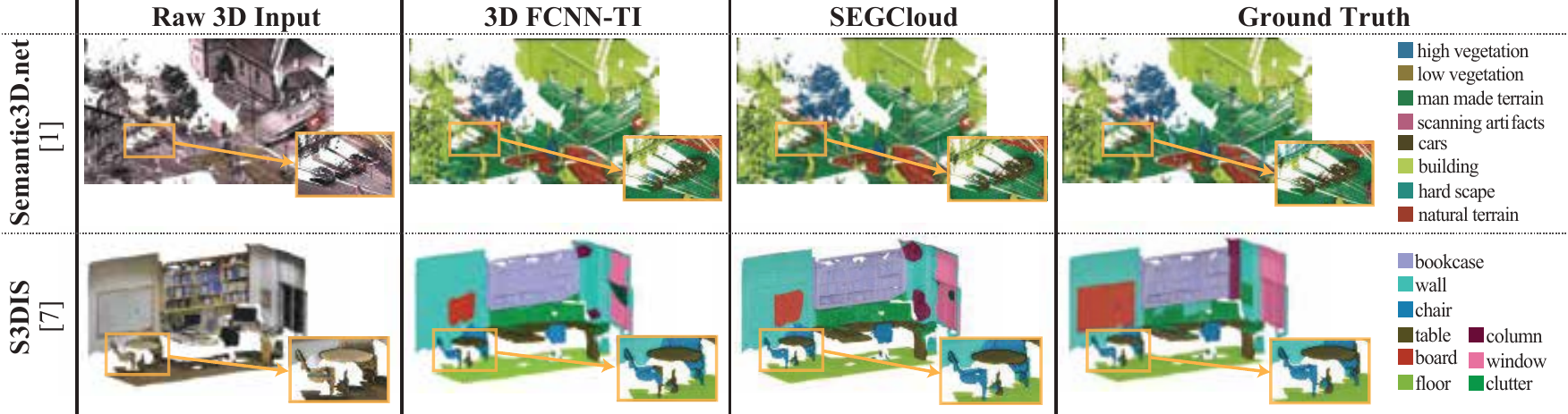}
\caption{\small{\textbf{Qualitative results of our framework on Semantic3D.net and S3DIS.} Additional results provided in suppl. \cite{segcloudsuppl}.}}
\label{fig:qualitative}
\end{figure*}

\textbf{Semantic3D.net benchmark:}
We evaluate our architecture on the recent Semantic3D.net benchmark~\cite{l3d}, which is currently the largest labeled 3D point cloud dataset for outdoor scenes. It contains over $3$ billion points and covers a range of urban scenes. We provide results on the \textit{reduced-8} challenge of the benchmark in Table~\ref{tab:large_scale_outdoors}. Our method outperforms ~\cite{snapnet} by 2.2 mIOU points and 2.28\% accuracy and sets a new state-of-the-art on that challenge. When compared against the best method that does not leverage extra data through ImageNet~\cite{ILSVRC15} pretrained networks, our method outperforms~\cite{Hackel2016} by 7.1 mIOU points, 4.1\% accuracy. Note that we also do not leverage extra data or ImageNet~\cite{ILSVRC15} pretrained networks. Our base \threedfcnn trained with Trilinear Interpolation (\threedfcnn-TI) already achieves state-of-the-art performance, and an additional improvement of 3.1 mIOU points and 3.22\% can be attributed to the \fccrf. An example segmentation of our method is shown in Figure \ref{fig:qualitative}. The \threedfcnn-TI produces a segmentation which contains some noise on the cars highlighted in the figure. However, the combination with the \fccrf in the \ours is able to remove the noise and provide a cleaner segmentation of the point cloud.\\

\textbf{Stanford Large-Scale 3D Indoor Spaces Dataset (S3DIS): }The S3DIS dataset~\cite{Armeni2016} provides 3D point clouds for six fully reconstructed large-scale areas, originating from three different buildings. We train our architecture on two of the buildings and test on the third. We compare our method against the MLP architecture of Qi~\etal, (PointNet)~\cite{DBLP:journals/corr/QiSMG16}. Qi~\etal.~\cite{DBLP:journals/corr/QiSMG16} perform a six-fold cross validation across areas rather than buildings. However, with this experimental setup, areas from the same building end up in both the training and test set resulting in increased performance and do not measure generalizability. For a more principled evaluation, we choose our test set to match their fifth fold (ie. we test on Area 5 and train on the rest). We obtain the results from the authors for comparison shown in Table~\ref{tab:S3DIS}. We outperform the MLP architeture of \cite{DBLP:journals/corr/QiSMG16} by 7.83 mIOU points and 8.37\% in mean accuracy. Our base \threedfcnn-TI also outperforms their architecture and the effect of our system's design choices on the performance of the \threedfcnn and \threedfcnn-TI are analyzed in Section \ref{sec:system}. Qualitative results on this dataset (Figure \ref{fig:qualitative}) show an example of how detailed boundaries are captured and refined by our method. \\

\textbf{NYU V2: }The NYU V2 dataset~\cite{Silberman:ECCV12} contains 1149 labeled RGB-D images. Camera parameters are available and are used to obtain a 3D point cloud for each RGB-D frame. In robotics and navigation applications, agents do not have access to fully reconstructed scenes and labeling single frame 3D point clouds becomes invaluable. We compare against 2D and 3D-based methods except for those that leverage additional large scale image datasets (\eg \cite{Kim2013}, \cite{eigen2015}), or do not use the official split or the 13-class labeling defined in \cite{couprie2013} (\eg \cite{Kim2013}, \cite{Wolf2015}). We obtain a confusion matrix for the highest performing method of~\cite{Wolf2016} to compute mean IOU in addition to the mean accuracy numbers they report. Wolf~\etal~\cite{Wolf2016} evaluate their method by aggregating results of 10 random forests. Similarly, we use 10 different random initializations of network weights, and use a validation set to select our final trained model for evaluation. Results are shown in Table \ref{tab:NYUV2}. We outperform the 3D Entangled Forests method of \cite{Wolf2016} by 3.94 mIOU points and 0.83\% mean accuracy. \\

\textbf{KITTI: }The KITTI dataset~\cite{Geiger2013IJRR,Geiger:2012:WRA:2354409.2354978} provides 6 hours of traffic recording using various sensors including a 3D laser scanner. Zhang~\etal.~\cite{Zhang2015} annotated a subset of the KITTI tracking dataset with 3D point cloud and corresponding 2D image annotations for use in sensor fusion for 2D semantic segmentation. As part of their sensor fusion process, they train a unimodal 3D point cloud classifier using Random Forests. 
We use this classifier as a baseline for evaluating our framework\textquotesingle s performance. The comparison on the labeled KITTI subset is reported in Table~\ref{tab:kitti}. We demonstrate performance on par with ~\cite{Zhang2015} where a Random Forests classifier is used for segmentation. Note that for this dataset, we train on the laser point cloud with no RGB information.\\

\textbf{Analysis of results: } In all datasets presented, our performance is on par with or better than previous methods. As expected, we also observe that the addition of a \fccrf improves the \threedfcnn-TI output and the qualitative results showcase its ability to recover clear object boundaries by smoothing out incorrect regions in the bilateral space (\eg. cars in Semantic3D.net or chairs in S3DIS). Quantitatively, it offers a relative  improvement of 3.0-5.3\% mIOU and 4.4-4.7\% mAcc for all datasets. Specifically, we see the largest relative improvement on Semantic3D.net - 5.3\% mIOU. Since Semantic3D.net is by far the largest dataset (at least 8X times larger), we believe that such characteristic might be representative for large scale datasets as the base networks are less prone to overfitting. We notice however that several classes in the S3DIS dataset, such as \textit{board}, \textit{column} and \textit{beam} are often incorrectly classified as \textit{walls}. These elements are often found in close proximity to walls and have similar colors, which can present a challenge to both the \threedfcnn-TI and the \fccrf. 

\subsection{System Analysis} \label{sec:system}
We analyze two additional components of our framework: geometric data augmentation and trilinear interpolation. The experiments presented in this section are performed on the S3DIS dataset. We also analyzed the effect of joint training versus separate CRF initialization (details and results in supplementary material \cite{segcloudsuppl}).\\ 
\begin{table*}
    \centering
    \resizebox{\textwidth}{!}{
    \begin{minipage}{0.50\textwidth}
        \centering
        \caption{\small{Effect of Geometric Augmentation}}
        \begin{tabular}[t]{r|c}
            \small{\textbf{Method}} & \small{\textbf{mIOU}} \\ \hline
            PointNet \cite{DBLP:journals/corr/QiSMG16} &  41.09 \\
            Ours- no augm. (\threedfcnn-TI) & 43.67 \\
            Ours (\threedfcnn-TI)  & \textbf{47.46} 
        \end{tabular}
        \label{tab:geo}    
    \end{minipage}
    \hfill

    \begin{minipage}{0.50\textwidth}
        \centering
        \caption{\small{Effect of trilinear interpolation}}            
        \begin{tabular}[t]{r|c}
            \small{\textbf{Method}} & \small{\textbf{mIOU}} \\ \hline
            PointNet \cite{DBLP:journals/corr/QiSMG16} & 41.09 \\
            Ours-NN (\threedfcnn-NN) & 44.84\\
            Ours (\threedfcnn-TI) & \textbf{47.46} 
        \end{tabular}
        \label{tab:trilinear}
    \end{minipage}
    \hfill
    }
    \vspace{-4mm}
\end{table*}

\textbf{Effect of Geometric Data Augmentation:} Our framework uses several types of data augmentation. Our geometric data augmentation methods in particular (random $360^{\circ}$ rotation along the z-axis and scaling) are non-standard. Qi~\etal. \cite{DBLP:journals/corr/QiSMG16} use  different augmentation, including random rotation along the z-axis, and jittering of $x, y, z$ coordinates to augment object 3D point clouds, but it is not specified whether the same augmentation is used on 3D scenes. We want to determine the role of our proposed geometric augmentation methods on the performance of our base \threedfcnn-TI architecture. We therefore train the \threedfcnn-TI without any geometric augmentation and report the performance in Table~\ref{tab:geo}. We observe that the geometric augmentation does play a significant role in the final performance and is responsible for an improvement of 3.79 mIOU points. However, even without any geometric augmentation, our base \threedfcnn-TI outperforms the MLP architecture of \cite{DBLP:journals/corr/QiSMG16} by 2.58 mIOU points.\\

\begin{figure}[t]
\centering
\includegraphics[width=.80\columnwidth]{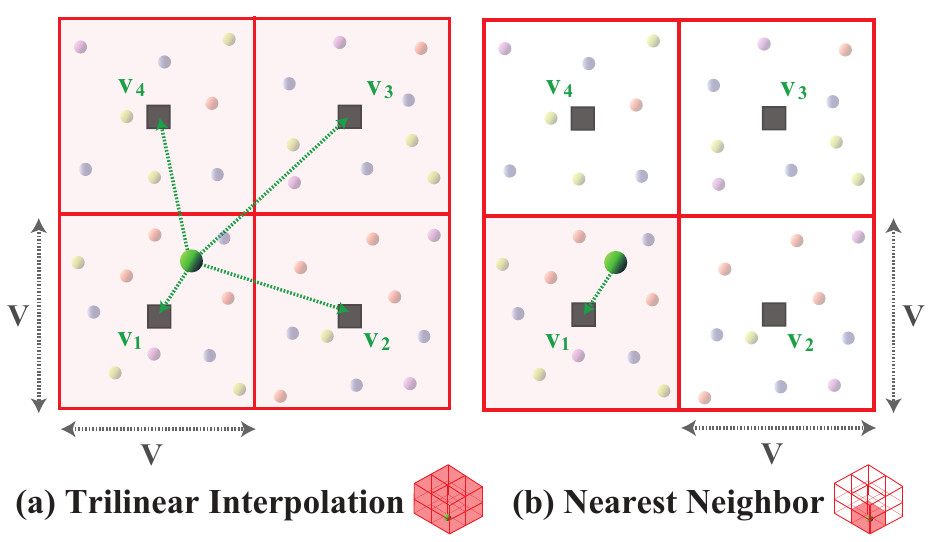}
\captionsetup{justification=centering}
\caption{\small{\textbf{Assigning voxel labels to 3D points (top view)}: Trilinear interpolation (a) versus the conventional approach of the nearest voxel center (b).}}
\label{fig:2DInterpol}
\vspace{-4mm}
\end{figure}

\textbf{Trilinear interpolation analysis:} We now present a study on the effect of trilinear interpolation on our framework. For simplicity, we perform this analysis on the combination of \threedfcnn and interpolation layer only (no \crf module). We want to study the advantage of our proposed 8-neighbours trilinear interpolation scheme (Section \ref{sec:trilinear}) over simply assigning labels of points according to the voxel they belong to (see Figure~\ref{fig:2DInterpol} for a schematic explanation of the two methods). The results of the two interpolation schemes are shown in Table \ref{tab:trilinear}. We observe that trilinear interpolation helps improve the \threedfcnn performance by 2.62 mIOU points over simply transferring the voxel label to the points within the voxel. This shows that considering the metric distance between points and voxels, as well a larger neighborhood of voxels can help improve accuracy in predictions.

\section{Conclusion}\label{sec:conclusion}
We presented an end-to-end 3D Semantic Segmentation framework that combines \threedfcnn, trilinear interpolation and \fccrf to provide class labels for 3D point clouds. Our approach achieves performance on par or better than state-of-the-art methods based on neural networks, randoms forests and graphical models. We show that several of its components such as geometric 3D data augmentation and trilinear interpolation play a key role in the final performance.
Although we demonstrate a clear advantage over some Random Forests methods and a point-based MLP method, our implementation uses a standard voxel-based \threedfcnn and could still adapt to the sparsity of the voxel grid using sparse convolutions (\eg \cite{octnet}) which could add an extra boost in performance, and set a new state-of-the-art in 3D Semantic Segmentation. \\

\textbf{Acknowledgments}

We acknowledge the support of Facebook and MURI (1186514-1-TBCJE) for this research.

\vspace{3mm}
{\small
\bibliographystyle{ieee}
\bibliography{egbib}
}

\ifarxiv
\clearpage
\appendix
\begin{strip}%
 \centering
 \LARGE \textbf{Appendix}
 \normalsize
\end{strip}

This document presents additional details and qualitative results for the framework presented in our main paper. Section \ref{sec:appdx_implementation} reports the particulars of our framework's implementation. The following section~\ref{sec:crf_init} offers details and results on the effect of using end-to-end training versus separate CRF intialization. The remaining of the document focuses on additional aspects of the evaluation and experiments. The experimental setup is detailed in Section \ref{sec:setup}. The characteristics of the datasets used in our evaluation are outlined in Section \ref{sec:appdx_datasets}. Section \ref{sec:metrics} defines the metrics used in evaluating our framework. Finally, qualitative results of our framework on all four datasets are illustrated in Section \ref{sec:visualizations}.

\section{Implementation}\label{sec:appdx_implementation}
This section provides additional implementation details, including procedures for 3D data augmentation, data preparation, training, as well as the programming framework.

\subsection{Augmentation Procedures for 3D data}
\label{sec:augmentation}

Most of the datasets we used are small to medium in scale. To make up for the lack of data, we perform a series of augmentations for 3D data. We apply the following data augmentations on-the-fly to increase randomness in the data and save storage space.

\textbf{Color Augmentation:} Color augmentation is a popular data augmentation technique for image datasets. We leverage it in our work by randomly varying the R, G and B channels of each observation within the range $\pm2.5$ for each channel.

\textbf{Geometric augmentation:} We also leverage 2 simple geometric augmentations: \textit{random rotation} and \textit{scaling}. We randomly rotate 3D observations around the axis along the gravity direction to mimic a change of viewpoints in a scene. During training, we sample rotation angles in the continuous range of $[0^{\circ}, 360^{\circ}]$ and rotate the point cloud on-the-fly. We also scale the data by a small factor that is uniformly sampled in the range $[0.9, 1.1]$ to make the network invariant to small changes in scale.

\textbf{Points Subsampling:} We also use a random sub-sampling of points in highly dense datasets, specifically, the Stanford Large-Scale 3D Indoor Spaces Dataset (S3DIS)~\cite{Armeni2016} and the Semantic3D.net~\cite{l3d}. During training, we sample points in a scene by a factor empirically chosen based on the number of points in the given point cloud crop (see Table \ref{tab:dsfactor}). For point clouds having more than $1e^5$ points, the sub-sampling factor for S3DIS is kept at 10 since the density of the point cloud is relatively constant in this dataset. The Semantic3D.net dataset on the other hand has varying density and we use three values of the sub-sampling factor (10, 50 and 100), as shown in Table \ref{tab:dsfactor}. This sub-sampling process aims at building invariance to missing points, and increasing the speed of the training process. At test time, the algorithm is evaluated on all input points without sub-sampling.

The above random transformations and sub-sampling allow us to increase the effective size of each dataset and can help the network build invariance to rotation/viewpoint changes, as well as reduced and varying context.

\begin{table}[ht]
    \centering
	\caption{\textbf{Cloud Sub-sampling Factor (For training-only)}}
    \label{tab:dsfactor}
	\vspace{0.5em}
	\begin{tabular}{|c|c|c|c|}
	\hline
	\diagbox[width=10em]{\textbf{Dataset}}
    {\shortstack{\textbf{Threshold} \\ (\#points)}} & \small{\textbf{$1e^5$}} & \small{\textbf{$1e^6$}} & \small{\textbf{$1e^7$}}\\
	\hline
	S3DIS & 10 & 10 & 10 \\
	Semantic3D.net & 10 & 50 & 100\\
	\hline
    \end{tabular}
\end{table}

\subsection{Input Preparation}
The large scale 3D observations are split into areas of at most $5 m$ in the $X$, $Y$ and $Z$ dimensions, where $Z$ is the gravity axis. One notable exception is the S3DIS dataset, which provides fully reconstructed 3D point clouds of indoor buildings spaces. For this dataset, we limit the $X$ and $Y$ dimensions to $5 m$ like rest of the datasets, but keep the entire $Z$ extent, which allows to include both the ceiling and floor in every crop. During training, such $5 m$ cropped sub-area overlap with adjacent sub-areas by $0.5 m$. There is no overlap at test time in order to obtain a single prediction per point. Sub-areas are then voxelized with a $5 cm$ resolution to obtain a maximum input volume of $100\times100\times100$. This granularity provides a balance between memory requirements and an adequate representation of the 3D space without information loss. Each voxel has one to five associated channels that correspond to its binary occupancy ($1$-occupied, $0$-empty), RGB value normalized within the range $[0, 1]$, and sensor intensity when available (Semantic3D.net dataset). The sensor intensity is mean centered and normalized using the mean and range of the training data distribution.

\subsection{Training}
\label{sec:training}

Training is performed in a 2-step process similar to~\cite{Zheng2015}. This process is illustrated in Figure \ref{fig:training_stages}. In the first training stage, we use the Trilinear Interpolation layer to map the voxel-wise predictions to point-wise predictions and minimize the point-wise loss. We train \threedfcnn with Trilinear Interpolation layer for $200$ epochs with a learning rate between $1e^-5$ and $1e^-3$, and reduce it by a factor of $10$ every $50$ epochs. In the second training stage, we combine the pre-trained \threedfcnn, the Trilinear Interpolation layer and the \crf, and train the whole system end-to-end. The base learning rate in this stage is set to a value between $1e^-7$ and $1e^-5$, and the training is performed for $2$ epochs. We use a learning rate multiplier of $1e^4$ and $1e^3$ for the \crf’s bilateral weights and compatibility matrix, however we did not extensively study the effect of these parameters. In most cases, the training of the second stage converges within a few hundred iterations (Convergence is determined using a validation set). In the CRF formulation, although the kernel weights $w_s, w_b$ and the compatibility matrix $\mu$ are learned using gradient descent, the kernel bandwidth parameters $\theta_\alpha, \theta_\beta, \theta_\gamma$ are not learned within our efficient variational inference framework. Thus, we used grid search or fixed values for some parameters following \cite{denseCRF}. We fix $\theta_\gamma$ at 5cm, $\theta_\beta$ at 11, and use a validation set to search for an optimal value of  $\theta_\alpha$. We limit our search to the range $[0.1, 3.2]m$. When no RGB information is available, we instead searched for $\theta_\gamma$  in the same range and did not use the bilateral filter. The kernel weights and compatibility matrix are learned during training. Similar to~\cite{Zheng2015} we use 5 CRF iterations during training and 10 CRF iterations at test time.

\begin{figure}[t]
\centering
\includegraphics[width=.90\linewidth]{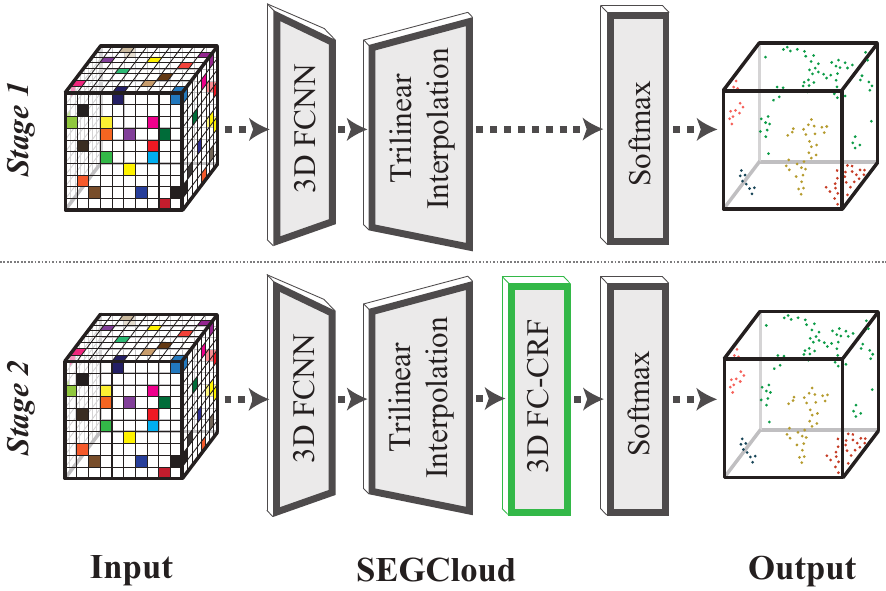}
\caption{\small{We follow a 2-stage training by first optimizing over the point-level unary potentials (no CRF) and then over the joint framework for point-level fine-grained labeling.}}
\label{fig:training_stages}
\vspace{-2mm}
\end{figure}

\begin{table*}[ht]
    \centering
	\caption{\textbf{Datasets Characteristics}}
	\vspace{5mm}
	\begin{tabular}{|c|c|c|c|c|}
	   & \small{KITTI~\cite{Geiger2013IJRR,Geiger:2012:WRA:2354409.2354978}} & \small{NYU V2~\cite{Silberman:ECCV12}} &  \small{S3DIS~\cite{Armeni2016}} & \small{Semantic3D.net~\cite{l3d}}\\
	 \hline
	Scene  & outdoor & indoor & indoor & outdoor\\
	Point Cloud type & partial & partial & full & partial\\
	Sensor type & Laser & Kinect & MatterPort & Laser\\

	Number of training points & 12million & 125million & 228million & 1.9billion\\	
    \end{tabular}
    \label{tab:datasets}
    \vspace{3mm}
\end{table*}

\section{Effect of end-to-end training vs separate CRF initialization} \label{sec:crf_init}
We performed an experiment to evaluate the effect of end-to-end training versus separately initializing the CRF module. For the separate initialization, we set the theta parameters to the optimal joint training values we found during end-to-end training, the spatial weight to 3, and the bilateral to 5 for all experiments. Results show that joint training performs better than separate CRF initialization especially in mAcc metric (see Table~\ref{tab:manualCRF}).
{\small
\tabcolsep= 1mm
\begin{table}[ht]
    \centering
    \caption{\small{\textbf{Effect of CRF initialization:} End-to-end training vs Manual}}
    \begin{tabular}{r|cc|cc}
        \multirow{2}{*}{\small{\textbf{Dataset}}} & \multicolumn{2}{|c}{\small{\textbf{End-to-end}}} & \multicolumn{2}{|c}{\small{\textbf{manual}}} \\
         & mIOU & mAcc & mIOU & mAcc \\ \hline
        Semantic3D.net & \textbf{61.30} & \textbf{73.08} & 60.72 & 69.69 \\
        S3DIS & \textbf{48.92} & \textbf{57.35} & 47.09 & 53.6\\
        KITTI & \textbf{36.78} & \textbf{49.46} & 36.34 & 46.34 \\
        NYUV2 & \textbf{43.45} & \textbf{56.43} & 41.63 & 52.28\\
    \end{tabular}
    \label{tab:manualCRF}
\end{table}}

\vspace{5mm}
\section{Experimental and Evaluation Setup}\label{sec:setup}

\subsection{Datasets}\label{sec:appdx_datasets}
We now present the characteristics of the datasets we use to evaluate our framework. The datasets we chose for evaluation are Semantic3D.net~\cite{l3d}, the Stanford Large-Scale 3D Indoor Spaces Dataset (S3DIS)~\cite{Armeni2016}, KITTI~\cite{Geiger2013IJRR,Geiger:2012:WRA:2354409.2354978}, and NYU V2~\cite{Silberman:ECCV12}. As shown in Table \ref{tab:datasets}, our framework is general in that it can handle point clouds from various sources, both indoor and outdoor environments, as well as partial and fully reconstructed point clouds. Specifically, two of the datasets are collected from indoor environments and two from outdoor environments. They also cover a variety of data acquisition methods, including laser scanners (Semantic3D.net, KITTI), Kinect (NYU V2), and MatterPort (S3DIS). Moreover, the S3DIS is a fully reconstructed point cloud dataset, while NYU V2 provides point clouds extracted from a single frame RGB-D camera. The size of the training sets also vary from $12$ million training points for the KITTI dataset to $1.9$ billion training points for Semantic3D.net (excluding the validation set).

\vspace{5mm}
\subsection{Evaluation Metrics} \label{sec:metrics}
We use two main metrics for our evaluation: \textit{mean class accuracy (mAcc)} and \textit{mean class IOU (mIOU)}, where IOU is defined similarly to the Pascal segmentation convention. Accuracy per class is defined as:
\begin{equation} \label{eq:acc}
    acc_i = \frac{tp_i}{gt_i} = \frac{tp_i}{tp_i + fn_i},
\end{equation}
where $tp_i$ is the number of true positives of class $i$, $fn_i$ is the number of false negatives of class $i$ and $gt_i$ is the total number of ground-truth elements of class $i$. The mean class accuracy is then defined as:
\begin{equation}
    mAcc = \frac{1}{N}\sum_{i=1}^{N}acc_i,
\end{equation}
where N is the number of classes.

We define per class IOU following the Pascal convention as:
\begin{equation}
    IOU_i = \frac{tp_i}{gt_i + fp_i}  = \frac{tp_i}{tp_i + fn_i + fp_i},
\end{equation}
where $tp_i, gt_i, fn_i$ are defined as above, and $fp_i$ is the number of false positives of class $i$. Note that IOU is a more difficult metric than accuracy since it doesn't simply reward true positives, but also penalizes false positives. From the definition above, we obtain mean class IOU as:
\begin{equation}
    mIOU = \frac{1}{N}\sum_{i=1}^{N}IOU_i.
\end{equation}

\vspace{5mm}
\subsection{Visualizations}\label{sec:visualizations}
In this section, we include more qualitative segmentation results for all datasets. The results showcase the initial segmentation of the standalone \threedfcnn-TI followed by the final result of the \ours framework.

\begin{figure*}
\centering
\includegraphics[width=.99\textwidth]{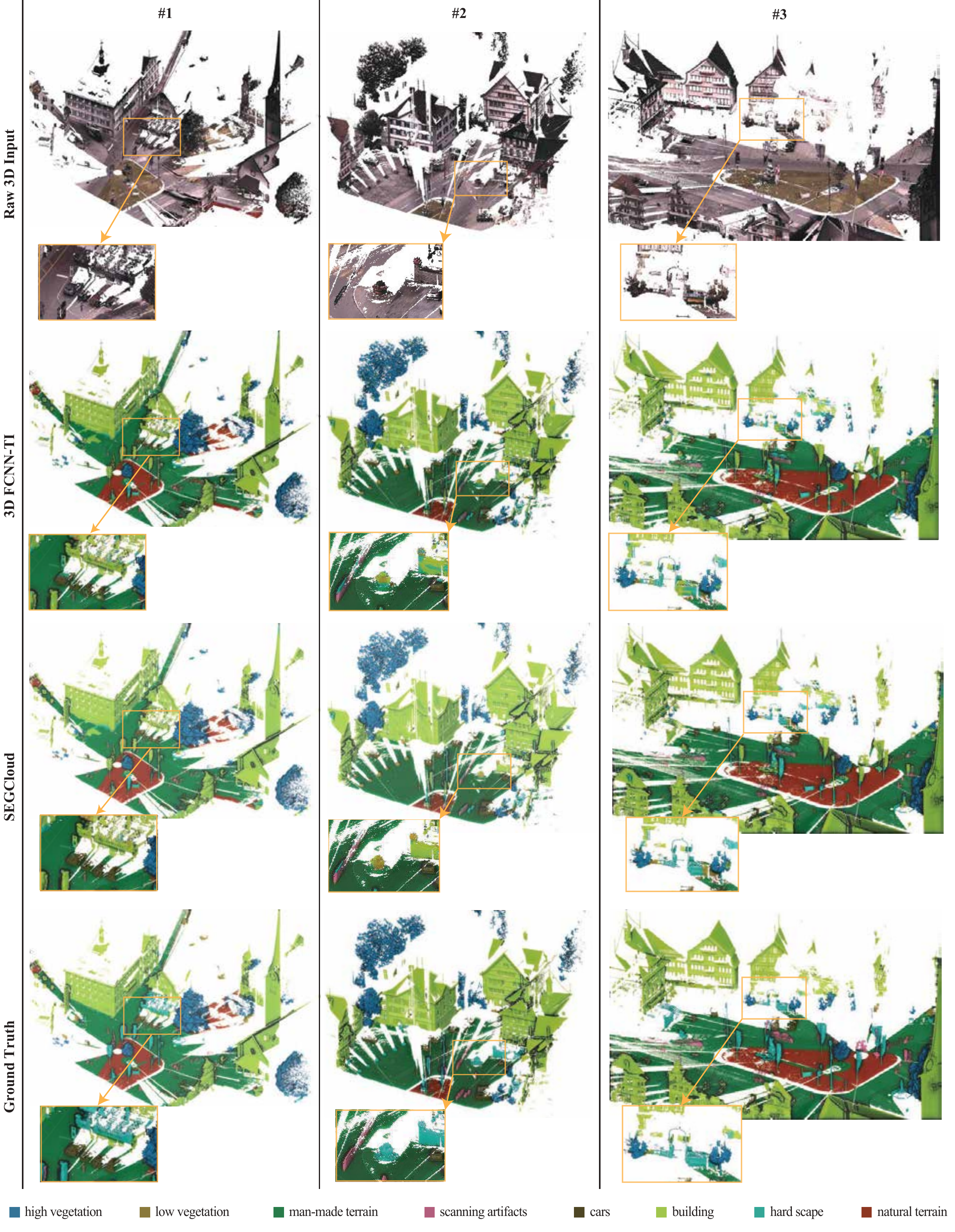}
\caption{\small{\textbf{Qualitative results on the Semantic3D.net dataset}}}
\label{fig:quals_l3d}
\end{figure*}

\begin{figure*}
\centering
\includegraphics[width=.90\textwidth]{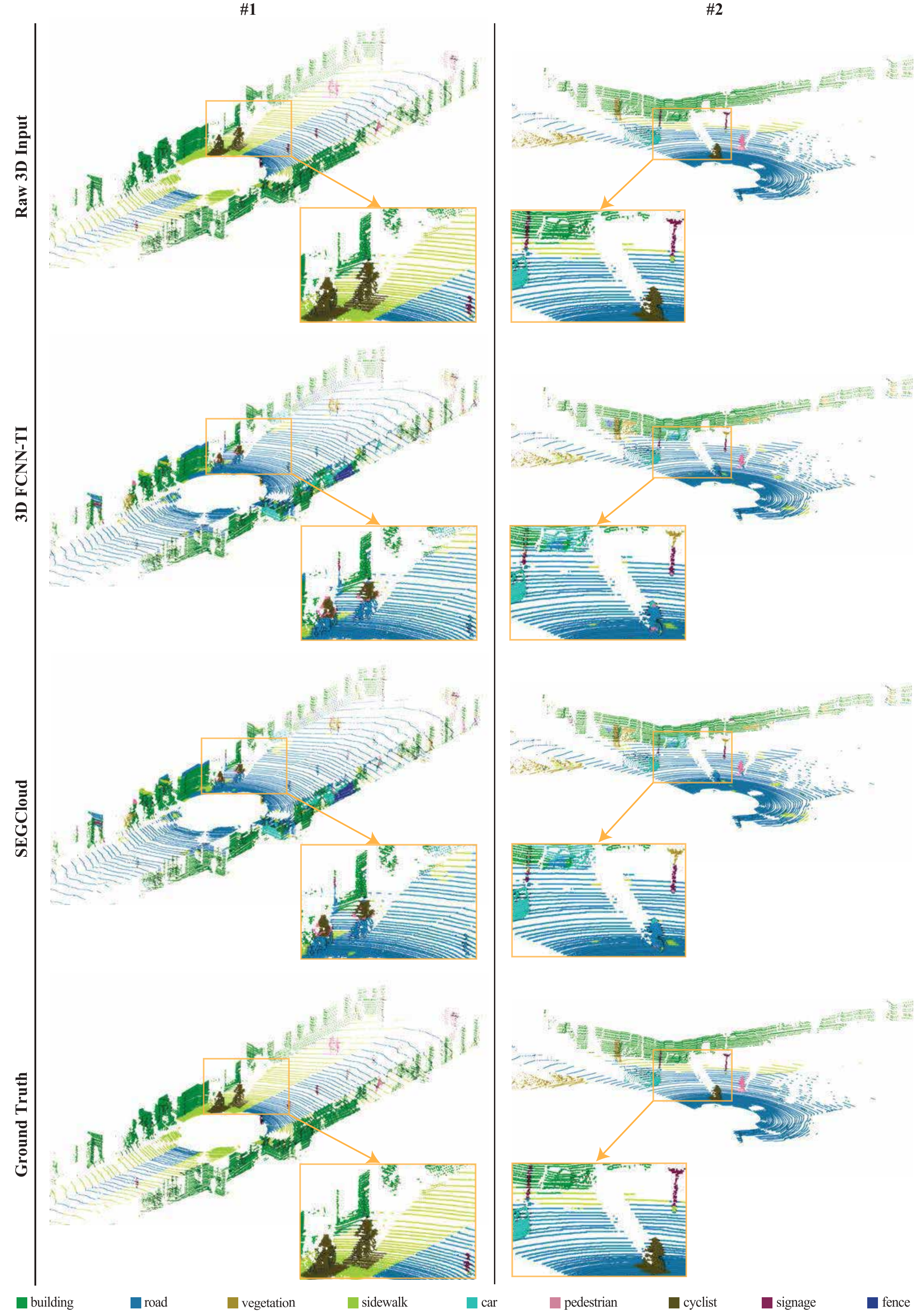}
\caption{\small{\textbf{Qualitative results on the KITTI dataset}}}
\label{fig:quals_kitti_1}
\end{figure*}

\begin{figure*}
\centering
\includegraphics[width=.90\textwidth]{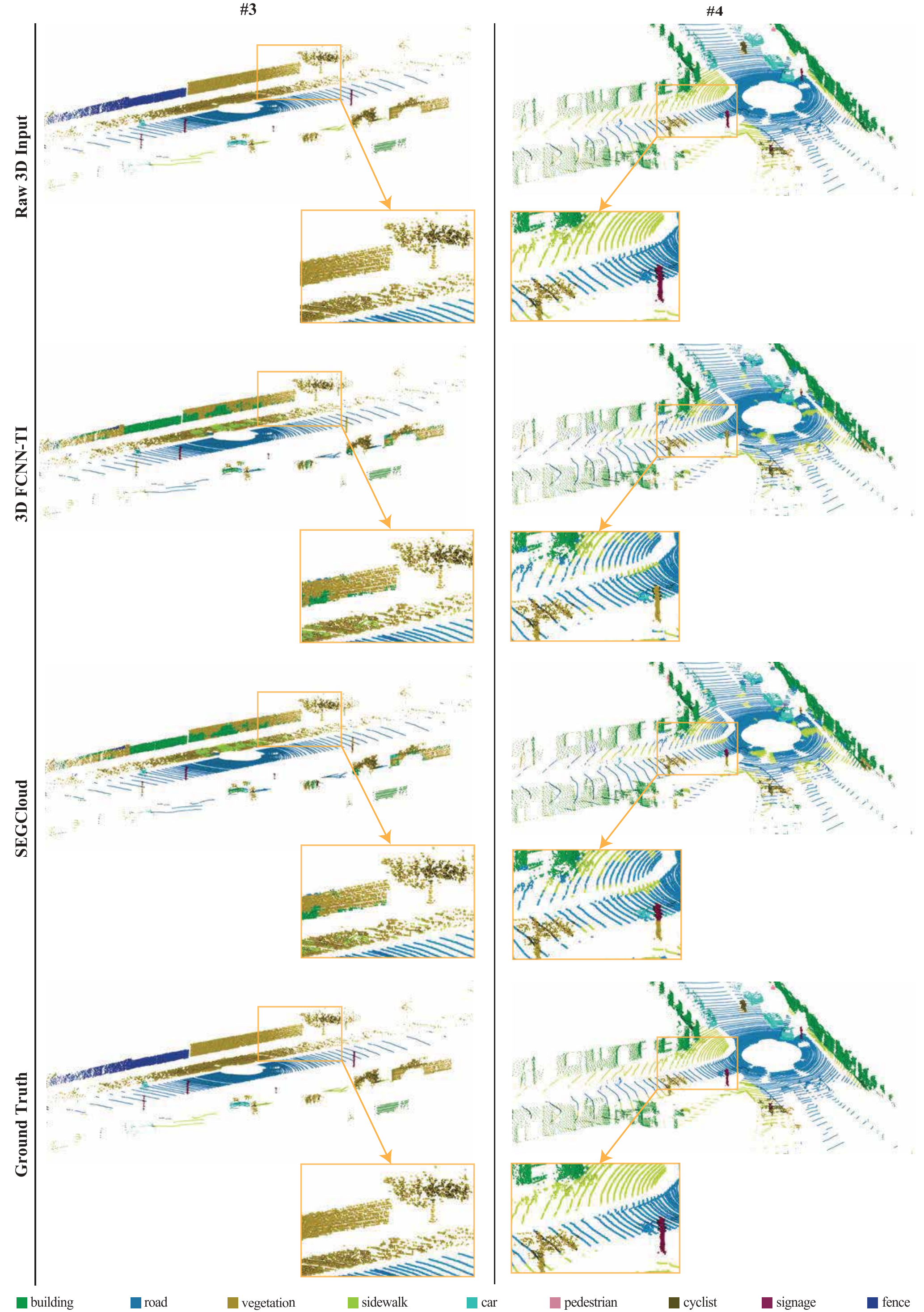}
\caption{\small{\textbf{Qualitative results on the KITTI dataset}}}
\label{fig:quals_kitti_2}
\end{figure*}

\begin{figure*}
\centering
\includegraphics[width=.77\textwidth]{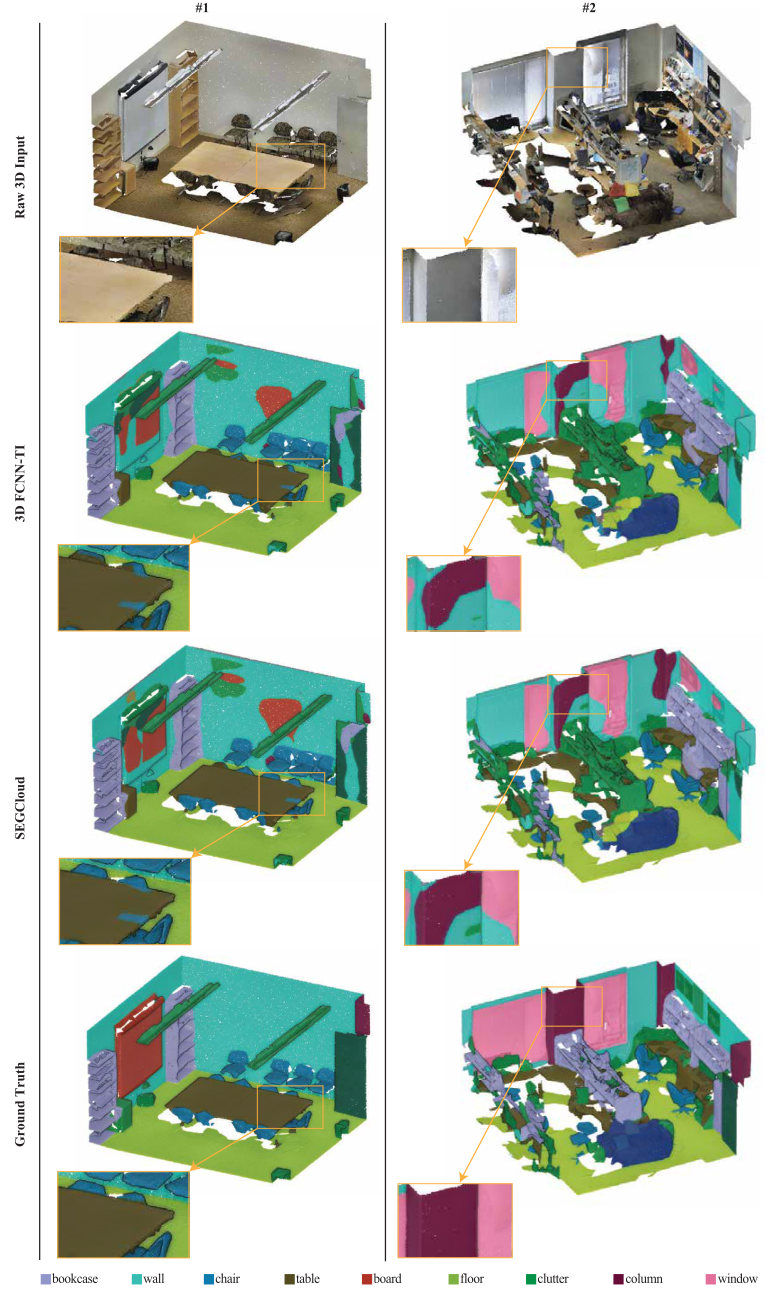}
\caption{\small{\textbf{Qualitative results on the S3DIS dataset}}}
\label{fig:quals_s3dis_1}
\end{figure*}

\begin{figure*}
\centering
\includegraphics[width=.77\textwidth]{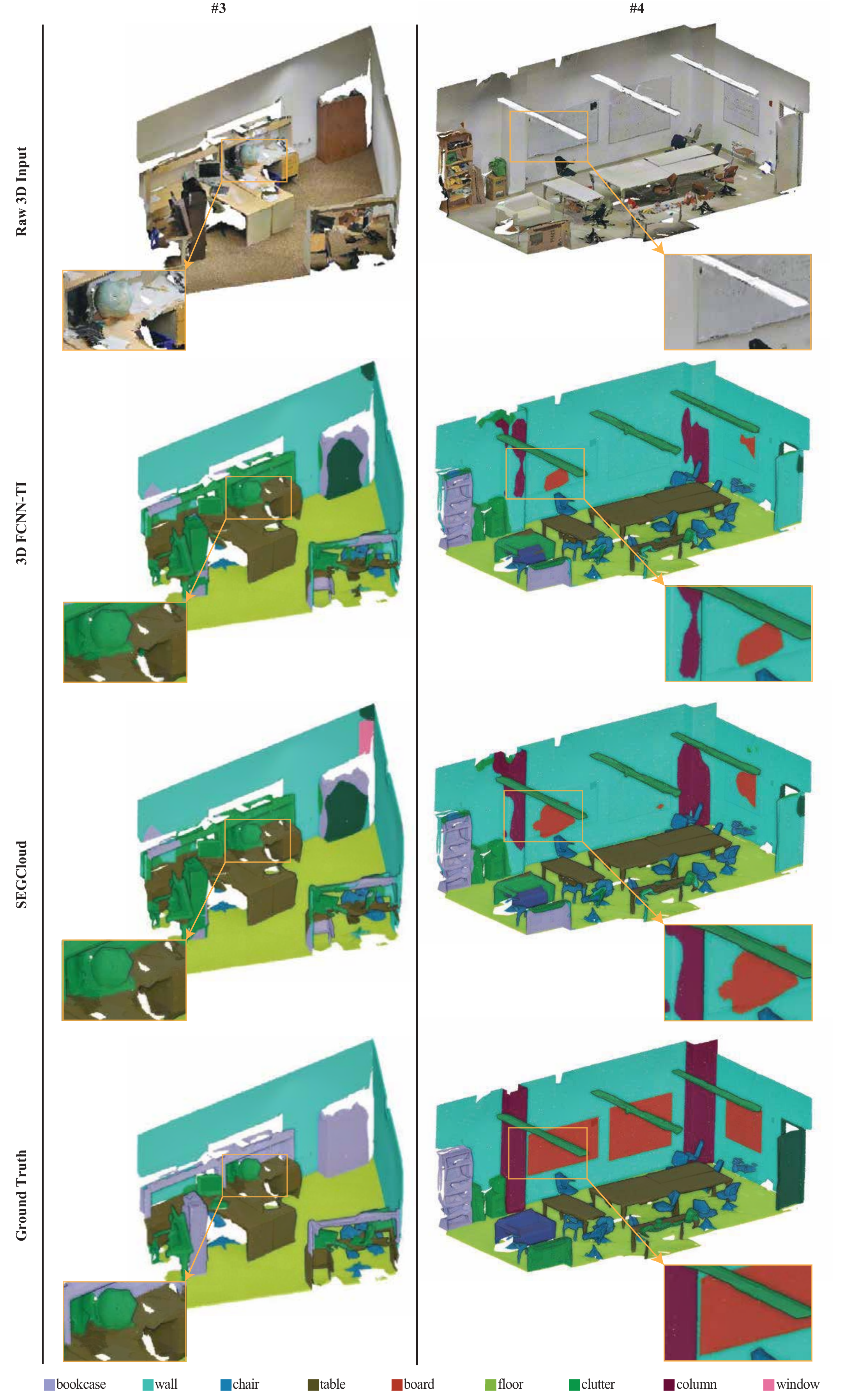}
\caption{\small{\textbf{Qualitative results on the S3DIS dataset}}}
\label{fig:quals_s3dis_2}
\end{figure*}

\begin{figure*}
\centering
\includegraphics[width=.99\textwidth]{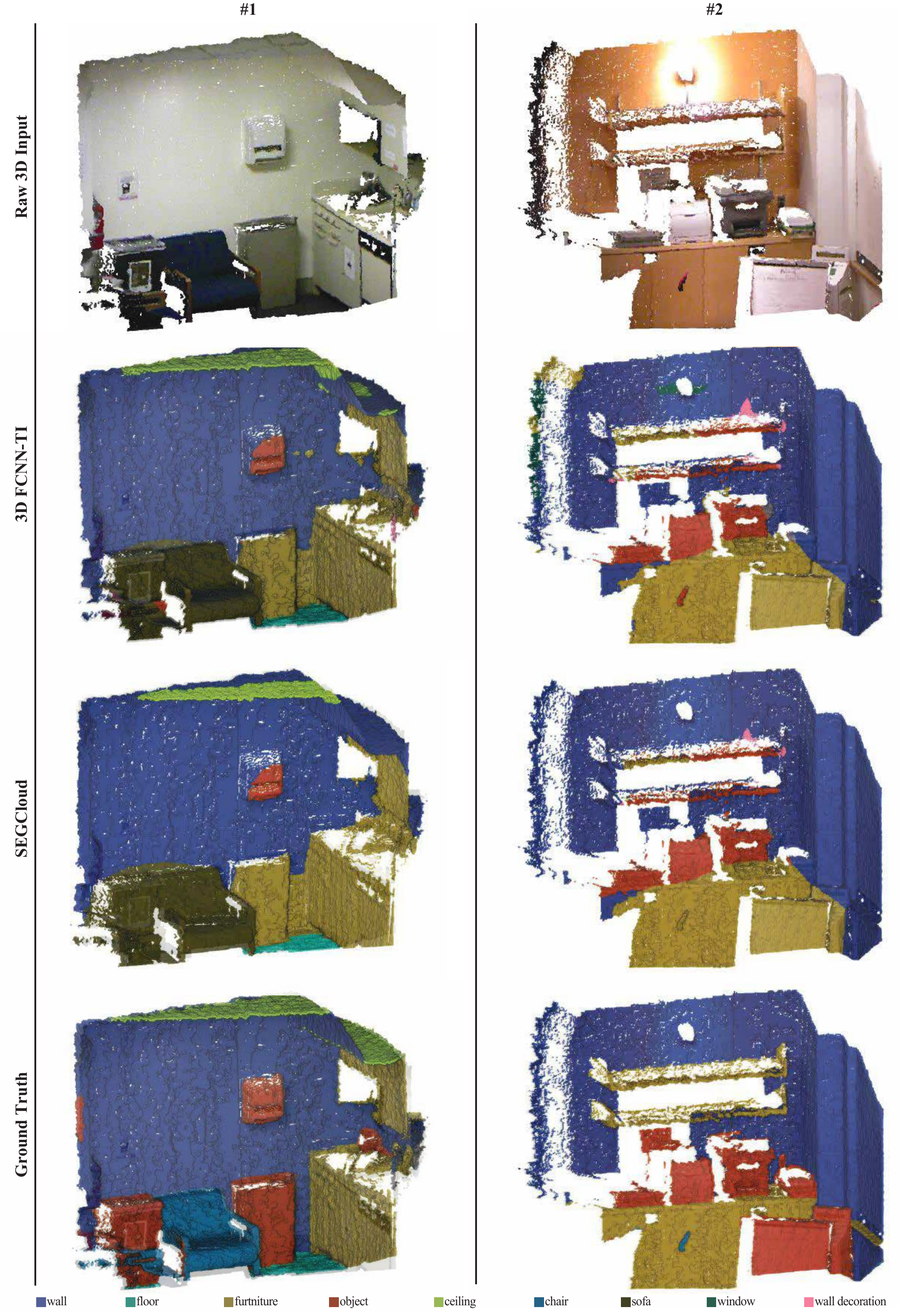}
\caption{\small{\textbf{Qualitative results on the NYU V2 dataset}}}
\label{fig:quals_nyuv2}
\end{figure*}

\begin{figure*}
\centering
\includegraphics[width=.99\textwidth]{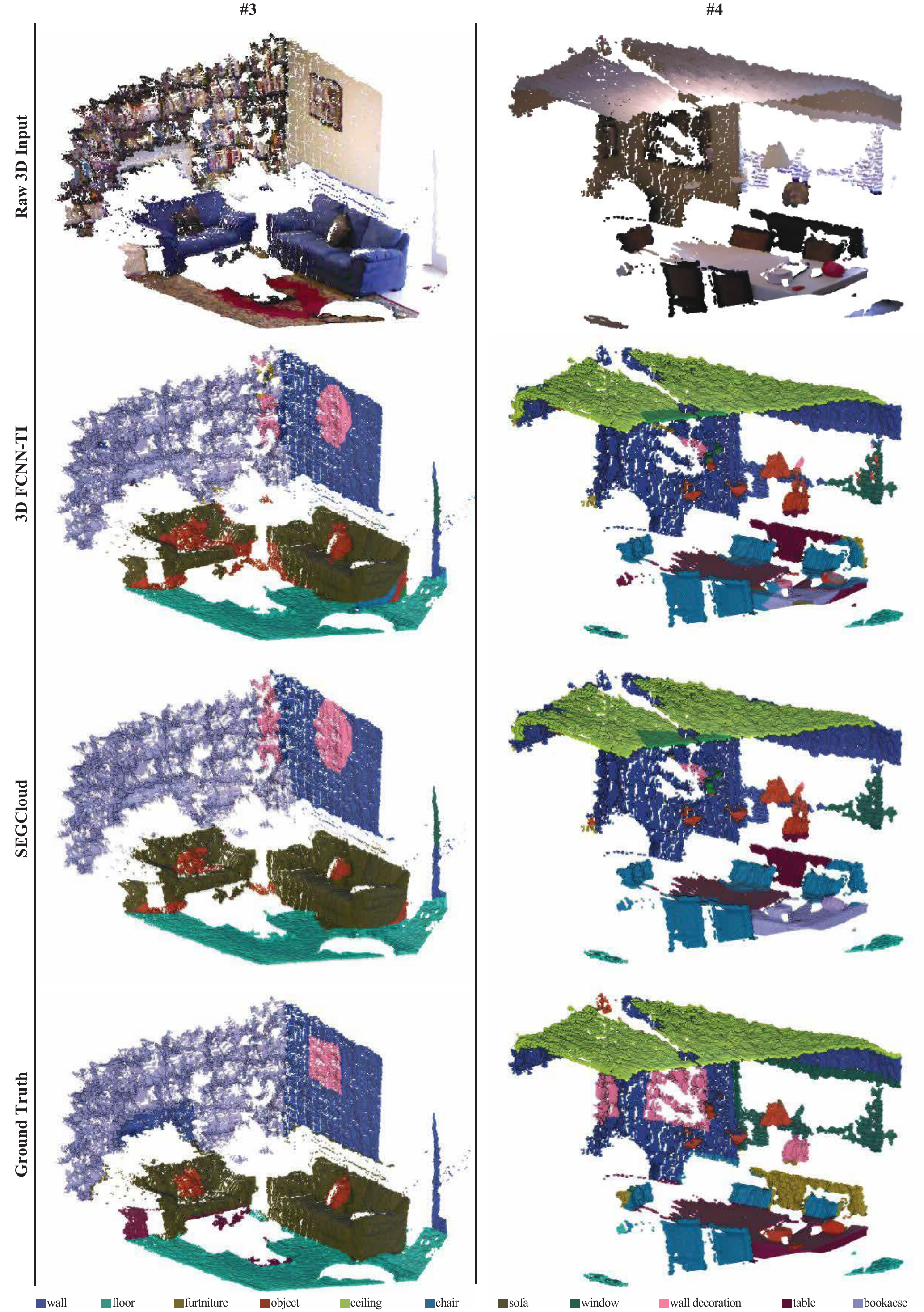}
\caption{\small{\textbf{Qualitative results on the NYU V2 dataset}}}
\label{fig:quals_nyuv2}
\end{figure*}
\fi

\end{document}